\documentclass{article}
\usepackage[T1]{fontenc}

\usepackage{microtype}
\usepackage{graphicx}
\usepackage{subfigure}
\usepackage{multirow}
\usepackage{amsmath}
\usepackage{amssymb}
\usepackage{placeins}
\usepackage{algorithm}
\usepackage{algorithmic}
\usepackage[table]{xcolor} 
\definecolor{lightgreen}{RGB}{235, 247, 238} 
\usepackage{booktabs} 

\usepackage{xurl}
\usepackage{hyperref}
\usepackage[accepted]{mlsys2025}
\usepackage{tcolorbox}

\definecolor{AQAbstractGray}{HTML}{F2F2F2}
\definecolor{AQRuleGray}{HTML}{909090}

\hypersetup{
  pdftitle={D-REAL: Acceptance-Length Optimization with Unbiased Gradients for Speculative Decoding},
  pdfauthor={Meng Wei, Yefei Chen, Antong Li, Yanan Gao, Xingliang Shi, Xin Gao, Zhiling Ye, Yun Yue, Chunxiao Guo, Peng Wei},
  pdfsubject={arXiv preprint},
  pdfkeywords={Speculative decoding, draft model training, expected acceptance length}
}

\begin{document}
\raggedbottom
\allowdisplaybreaks[1]

\pagestyle{plain}
\setlength{\footskip}{20pt}

\twocolumn[{
\noindent
\begin{minipage}{\textwidth}
\setlength{\parindent}{0pt}
\setlength{\parskip}{0pt}

\noindent
\begin{minipage}[c]{32mm} 
  \includegraphics[
    height=12mm 
  ]{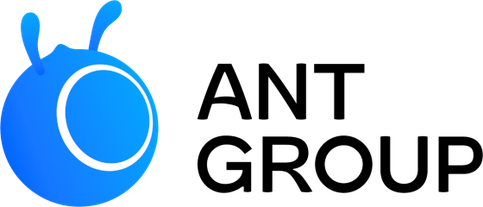}
\end{minipage}%
\hspace{0.5mm}
{\color{gray!50}\rule[-3.5mm]{0.6pt}{7mm}}
\hspace{6mm}
\begin{minipage}[c]{14mm} 
  \includegraphics[
    trim=64bp 64bp 40bp 44bp,
    clip,
    height=12mm 
  ]{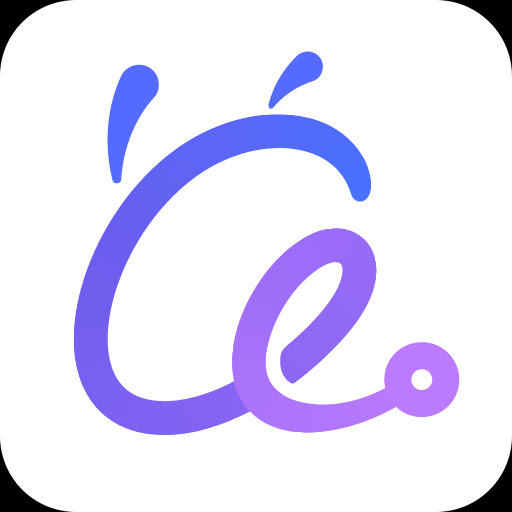}
\end{minipage}%
\begin{minipage}[c]{0.6\linewidth}
  {\fontencoding{T1}%
   \fontfamily{qag}\fontseries{m}\fontshape{n}%
   \fontsize{20}{19}\selectfont
   \textls[15]{AQ MedicalAI}\par}
\end{minipage}
\par

\vspace{6pt}
{\color{AQRuleGray}\rule{\linewidth}{0.5pt}\par}
\vspace{14pt}

{\raggedright
 \bfseries\fontsize{20}{24}\selectfont
 D-REAL: Acceptance-Length Optimization\\
 with Unbiased Gradients for Speculative Decoding\par}

\vspace{16pt}

{\raggedright
 \bfseries\fontsize{11}{15}\selectfont
 Meng Wei,\quad
 Yefei Chen,\quad
 Antong Li,\quad
 Yanan Gao,\quad
 Xingliang Shi,\\
 Xin Gao,\quad
 Zhiling Ye,\quad
 Yun Yue,\quad
 Chunxiao Guo,\quad
 Peng Wei\textsuperscript{\(\dagger\)}
 \par}

\vspace{5pt}

\vspace{4pt}

{\fontsize{9}{12}\selectfont
 \textsuperscript{\(\dagger\)}Corresponding author
 \par}

\vspace{13pt}

\begin{tcolorbox}[
  width=\linewidth,
  colback=AQAbstractGray,
  colframe=AQAbstractGray,
  boxrule=0pt,
  arc=3mm,
  boxsep=0pt,
  left=4mm,
  right=4mm,
  top=3mm,
  bottom=3mm,
  before skip=0pt,
  after skip=0pt,
  fontupper=\fontsize{10.5}{13.2}\selectfont
]
Speculative decoding efficiency depends on the number of
consecutively accepted draft tokens, yet token-level training
objectives do not directly optimize this quantity.
We introduce \textbf{D-REAL}
(\textbf{D}rafter \textbf{R}einforcement of
\textbf{E}xpected \textbf{A}cceptance \textbf{L}ength),
a post-training method that directly optimizes expected
acceptance length for a fixed decoding context.
D-REAL weights position-wise draft--target distributional
overlap by prefix survival probabilities along a draft-sampled
prefix, faithfully capturing the sequential dependence of
verification on earlier proposals.
By accounting for the parameter dependence of the proposal
distribution, we derive an unbiased gradient estimator for
this single-round objective under stochastic verification.
We evaluate D-REAL on DSpark and DFlash2 with Qwen3-8B as
the target model across ten benchmarks.
Starting from baselines whose mean acceptance length has
plateaued, one epoch of D-REAL post-training consistently
improves the benchmark-averaged mean acceptance length by
up to 0.23 tokens across both architectures under greedy
($T=0.0$) and stochastic ($T=1.0$) sampling, translating
into up to a 9.4\% end-to-end serving throughput improvement.
\end{tcolorbox}

\end{minipage}
\par\vspace{14pt}
}]

\thispagestyle{empty}

\section{Introduction}
Speculative decoding reduces sequential target-model evaluations by letting a small drafter propose candidate tokens that a larger model verifies in parallel~\citep{leviathan2023fast}. With fixed drafting and verification overhead per step, the end-to-end speedup is primarily governed by the mean acceptance length---the number of draft tokens consecutively accepted by the target model. A rejection prevents later proposals in the same block from contributing to this prefix, making acceptance length a sequential training objective.

Standard token-level objectives such as cross-entropy and KL divergence do not explicitly optimize the length of the accepted draft prefix. D-PACE~\citep{wu2026dpace} addresses positional credit assignment using weights derived from an accepted-length proxy on reference-token confidences. Our objective differs by targeting the actual single-round expectation under stochastic rejection sampling. This requires evaluating acceptance on draft-sampled prefixes and accounting for the parameter dependence of their sampling distribution. Earlier proposals determine the conditioning context for subsequent draft predictions and target verification, including when a parallel backbone is followed by token-conditioned heads. A correct gradient estimator must therefore account for both the explicit derivatives of the acceptance factors and the contribution from the changing proposal distribution. 

To address this challenge, we introduce \textbf{D-REAL}, a post-training method that directly optimizes the single-round expected accepted draft length under stochastic rejection sampling. D-REAL evaluates target--draft distributional overlap at each position weighted by prefix survival probabilities along actual draft-sampled trajectories. By constructing a differentiable surrogate using importance-weighted acceptance ratios with stop-gradient denominators, D-REAL captures the gradient flowing through the evolving proposal distribution. 

Our main contributions are summarized as follows:
\begin{itemize}
    \item \textbf{Unbiased Acceptance-Length Gradient Estimation.}
    We formulate single-round speculative decoding training as expected acceptance-length maximization over fixed-context, draft-sampled prefixes and derive \textbf{D-REAL}, a closed-form surrogate loss that yields an unbiased gradient estimator under local support and differentiability conditions (Section~\ref{sec:unbiased_acceptance}).

    \item \textbf{Unifying Distributional Overlap and Policy Gradients.}
    By pairing online total-variation overlap with importance-weighted survival ratios using stop-gradient proposal denominators, D-REAL naturally captures the score-function gradient of the proposal trajectory without heuristic baselines or clipping, which we efficiently realize via two-pass replay and prefix-shared online verification (Section~\ref{sec:implementation}).

    \item \textbf{Consistent Post-Training and Serving Gains.}
    Across ten benchmarks on DSpark and DFlash2, one epoch of D-REAL post-training consistently improves macro-average acceptance length ($\tau$) over both continued baseline training and D-PACE at $T=0.0$ and $T=1.0$, translating into up to $9.4\%$ higher serving throughput on GSM8K at concurrency $32$ (Section~\ref{sec:exp_results}).
\end{itemize}

\section{Related Work}
\label{sec:related_work}

\paragraph{Speculative Decoding and Parallel Drafters.}
Speculative decoding accelerates inference by decoupling candidate token generation from verification~\citep{leviathan2023fast, chen2023accelerating}. 
Early drafters predominantly relied on small autoregressive models~\citep{miao2024specinfer} or feature-level recurrent prediction heads such as the Medusa and EAGLE families~\citep{cai2024medusa, li2024eagle, li2024eagle2}. 
To break the sequential latency bottleneck within the drafter itself, recent approaches turn to non-autoregressive or semi-parallel architectures that predict multi-token blocks in a single forward pass, including block diffusion models~\citep{arriola2025block, li2026diffuspec} and parallel draft backbones like DFlash and DART~\citep{chen2026dflash, liu2026dart}.
While parallel drafters drastically reduce drafting latency by computing block-level representations concurrently, later token predictions and target verification still sequentially depend on earlier unverified candidates---making trajectory-aware training a fundamental challenge.

\paragraph{Positional Heuristics and Acceptance-Aware Objectives.}
Standard token-level objectives (e.g., cross-entropy and forward KL) use position-wise additive losses without explicitly optimizing accepted-prefix length. To capture the fact that later tokens are discarded if an earlier token is rejected, contemporary drafters introduce position-dependent schedules. 
Specifically, DFlash and DART employ a static exponential decay, $w_t = \exp(-(t-1)/\gamma)$, penalizing errors in later positions less heavily~\citep{chen2026dflash, liu2026dart}. 
Recognizing the limitations of static schedules, recent works seek acceptance-aware training objectives~\citep{wu2026dpal,xia2026acceptance,samarin2026lk}. 
GRIFFIN heuristically masks loss at position $t$ if any preceding target token falls outside the top-$K$ predictions~\citep{hu2025griffin}. 
More recently, D-PACE~\citep{wu2026dpace} derives position coefficients from a confidence-based accepted-length proxy and forms detached cross-entropy weights from smoothed draft confidences. The coefficients encode both prefix confidence and continuation value. Our comparison focuses on the distinction between this reference-prefix surrogate and the actual rejection-sampling expectation over draft-generated prefixes; detachment alone does not establish whether a gradient estimator is biased.

\paragraph{Distributional Matching and Policy Gradient Optimization.}
Under stochastic rejection sampling, the acceptance probability at a fixed prefix corresponds to the distribution overlap $1 - \operatorname{TV}(p, q) = \sum_{v \in \mathcal{V}} \min\{p(v), q(v)\}$~\citep{leviathan2023fast, zhou2024distillspec}.
Several target-supervision methods attempt to align the drafter with verification signals via online feedback, streak distillation, or reinforcement learning (e.g., DistillSpec, SpecDiff-2, and CORAL)~\citep{zhou2024distillspec, sandler2026specdiff2, weng2025coral}. 
For an objective averaged over draft-sampled trajectories, the gradient must account for both the acceptance probabilities and the parameter-dependent trajectory distribution. \textbf{D-REAL} exploits the product structure of accepted-prefix probabilities to incorporate these contributions through fixed behavior-proposal denominators. Its guarantee concerns the expected accepted draft length at a fixed context, as formalized in Section~\ref{sec:unbiased_acceptance}.

\section{D-REAL}
\label{sec:method}
\subsection{Problem Formulation}
\label{sec:preliminaries}

\paragraph{Speculative decoding and verification.}
Let $p$ denote a frozen target model and $q_\theta$ a smaller draft model parameterized by $\theta$, where both distributions subsume decoding temperatures. Given a committed context $c$, the drafter proposes a block of $K$ candidate tokens $D_{1:K}$ from $q_\theta$, which the target verifies in parallel in a single forward pass. For brevity, we denote the step-$t$ target and draft distributions conditioned on the prefix $D_{<t}$ as $p_t(v) := p(v \mid c, D_{<t})$ and $q_{\theta,t}(v) := q_\theta(v \mid c, D_{<t})$ for any token $v \in \mathcal{V}$. 

During verification, tokens are evaluated sequentially from left to right: at position $t$, candidate $D_t$ is accepted with probability
\begin{equation}
    \alpha_{\theta,t}(D_t) = \min\left\{1, \frac{p_t(D_t)}{q_{\theta,t}(D_t)}\right\}.
    \label{eq:acceptance_rule}
\end{equation}
Verification terminates at the first rejected token, after which a corrective token is drawn from the residual distribution to strictly recover the target distribution~\citep{leviathan2023fast}. 
The accepted length in a block is given by
\begin{equation}
\begin{aligned}
N=\max\bigl\{k\in\{0,\dots,K\}:\;&D_i\text{ is accepted}\\
 &\text{for all }i\le k\bigr\}.
\end{aligned}
\label{eq:accepted_length}
\end{equation}
Accounting for the additional bonus (or corrective) token guaranteed by the target verifier in each round, the standard \emph{mean acceptance length} per verification round is defined as
\begin{equation}
    \tau = \mathbb{E}[N] + 1,
    \label{eq:emitted_length}
\end{equation}
which serves as the primary determinant of speculative decoding speedup under fixed drafting and verification costs.

\paragraph{Drafter architectures and training objective.}
Our formulation applies generally to both autoregressive and parallel drafters. Modern parallel drafters such as DSpark~\citep{cheng2026dspark} and DFlash2~\citep{inco2026dflash2} decouple expensive feature extraction from token-level conditioning: their transformer backbones first compute block-wide hidden states $H_{1:K}(c)$ in a single parallel forward pass, after which a lightweight conditional head sequentially incorporates the embeddings of earlier sampled tokens $D_{<t}$ to parameterize $q_{\theta,t}(v) = q_\theta(v \mid H_t(c), D_{<t})$. Consequently, sequential dependencies arise from two coupled sources: both the draft distribution $q_{\theta,t}$ (via the lightweight head) and the target verification distribution $p_t(v) = p(v \mid c, D_{<t})$ at position $t$ explicitly condition on the realization of earlier draft proposals $D_{<t}$ drawn from the joint distribution $Q_\theta(D_{1:K}) = \prod_{j=1}^K q_{\theta,j}(D_j)$. Given training contexts $c \sim \rho$, maximizing $\tau$ is equivalent to maximizing the expected number of accepted draft tokens $\mathbb{E}[N]$:
\begin{equation}
\begin{aligned}
&\max_\theta\,\mathbb{E}_{c\sim\rho}[J(\theta;c)],\quad\text{where}\\
&J(\theta;c):=\mathbb{E}_{D_{1:K}\sim Q_\theta,\,\text{verify}}[N\mid c].
\end{aligned}
\label{eq:training_objective}
\end{equation}
Below, we derive an unbiased gradient estimator for $J(\theta; c)$ that preserves this dual sequential coupling and accounts for the parameter dependence of $Q_\theta$.

\begin{figure*}[t]
    \centering
    \includegraphics[width=\textwidth]{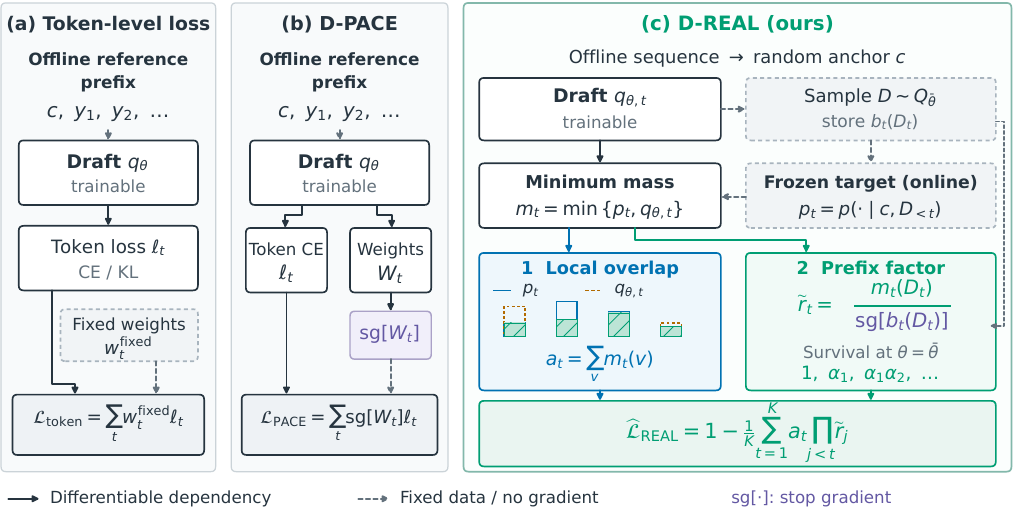}
    \caption{Overview of draft-training objectives. Unlike (a) standard token-level losses and (b) D-PACE, which weight token likelihoods along fixed offline reference prefixes, (c) \textbf{D-REAL} directly optimizes expected acceptance length along online draft-sampled trajectories by combining exact local distributional overlap (\emph{left inset}) with differentiable prefix survival factors (\emph{right inset}) via stop-gradient proposal denominators.}
\label{fig:dreal_overview}
\end{figure*}

\subsection{Unbiased Optimization of Expected Acceptance Length}
\label{sec:unbiased_acceptance}

\paragraph{An unbiased accepted-length estimator.}
To evaluate the expectation in Eq.~\eqref{eq:training_objective} for a given context $c$, we define the step-$t$ pointwise minimum mass and the total distributional overlap as:
\begin{align}
m_{\theta,t}(v) &= \min\{p_t(v), q_{\theta,t}(v)\}, \label{eq:dreal_mass}\\
a_{\theta,t} &= \sum_{v \in \mathcal{V}} m_{\theta,t}(v) = 1 - \operatorname{TV}(p_t, q_{\theta,t}), \label{eq:dreal_overlap}
\end{align}
where $\operatorname{TV}$ is the total variation distance. Note that $\alpha_{\theta,t}(v) = m_{\theta,t}(v) / q_{\theta,t}(v)$, and the overlap $a_{\theta,t} = \mathbb{E}_{D_t \sim q_{\theta,t}}[\alpha_{\theta,t}(D_t)]$ represents the expected acceptance probability of the $t$-th proposal conditioned on the prefix $D_{<t}$.
Applying the tail-sum formula to $J(\theta; c)$ in Eq.~\eqref{eq:training_objective}, we obtain:
\begin{align}
J(\theta;c)
&=\sum_{t=1}^{K}\Pr(N\geq t\mid c)\notag\\
&=\sum_{t=1}^{K}\mathbb{E}_{D_{\leq t}\sim Q_\theta}\!\left[\prod_{j=1}^{t}\alpha_{\theta,j}(D_j)\right]\notag\\
&= \mathbb{E}_{D\sim Q_\theta} \left[ \sum_{t=1}^{K}a_{\theta,t} \prod_{j<t}\alpha_{\theta,j}(D_j) \right], \label{eq:dreal_expected_length}
\end{align}
where the last equality follows from marginalizing out $D_t$ at each step $t$.
The term inside the expectation, denoted $\widehat J_\theta(D;c)$, is an unbiased estimator of $J(\theta;c)$. Notice that each overlap $a_{\theta,t}$ is weighted by the survival probability along the sampled prefix, naturally preserving inter-token dependencies.

\paragraph{Accounting for the proposal-distribution gradient.}
Figure~\ref{fig:dreal_overview} compares the three training objectives. While token-level training and D-PACE weight token losses along offline reference prefixes, D-REAL evaluates local distributional overlap along draft-sampled prefixes and weights it by prefix survival. The insets in Figure~\ref{fig:dreal_overview}(c) show the shared probability mass and the survival weights before successive draft positions.
At $\theta=\bar\theta$, these weights are $1$, $\alpha_1$, $\alpha_1\alpha_2$, and so on; their differentiable counterparts use $\widetilde r_j$ with fixed behavior-proposal denominators. To capture the true gradient of $J(\theta;c)$ over evolving draft trajectories, let $\bar\theta$ denote the parameter snapshot used to generate $D_{1:K}$, and write $b_t(v) := q_{\bar\theta,t}(v)$. Storing the proposal probabilities $b_t(D_t)$, we construct the differentiable surrogate:
\begin{align}
\widetilde r_{\theta,t} &= \frac{m_{\theta,t}(D_t)}{\operatorname{sg}[b_t(D_t)]}, \label{eq:dreal_ratio}\\
\widetilde J_{\theta;\bar\theta}(D;c) &= \sum_{t=1}^{K}a_{\theta,t} \prod_{j<t}\widetilde r_{\theta,j}, \label{eq:dreal_surrogate}\\
\widehat{\mathcal L}_{\mathrm{D\text{-}REAL}} &= 1 - \frac{1}{K}\widetilde J_{\theta;\bar\theta}(D;c), \label{eq:dreal_loss}
\end{align}
where $\operatorname{sg}[\cdot]$ is the stop-gradient operator. We evaluate both the loss and its gradient at $\theta = \bar\theta$, yielding $\widetilde r_{\bar\theta,t} = \alpha_{\bar\theta,t}(D_t)$. 

Assuming the proposal support is locally invariant around $\bar\theta$, and noting that $m_{\theta,t}$ is differentiable almost everywhere (assuming the tie-set $\{v : p_t(v) = q_{\theta,t}(v)\}$ has Lebesgue measure zero in the parameter space), we explicitly write the prefix dependence---$a_{\theta,t}(d_{<t})$, $m_{\theta,j}(d_j; d_{<j})$, and $b_j(d_j; d_{<j})$---and expand the expectation over all prefix realizations $d_{<t} \in \mathcal{V}^{t-1}$:
{\small
\begin{align}
&\mathbb{E}_{D\sim Q_{\bar\theta}}[\widetilde J_{\theta;\bar\theta}(D;c)]\notag\\
&\quad=\sum_{t=1}^{K}\sum_{d_{<t}}a_{\theta,t}(d_{<t})
\prod_{j<t}\!\left[b_j(d_j;d_{<j})\frac{m_{\theta,j}(d_j;d_{<j})}{b_j(d_j;d_{<j})}\right]\notag\\
&\quad=\sum_{t=1}^{K}\sum_{d_{<t}}a_{\theta,t}(d_{<t})\prod_{j<t}m_{\theta,j}(d_j;d_{<j})\notag\\
&\quad=\sum_{t=1}^{K}\sum_{d_{<t}}\prod_{j<t}q_{\theta,j}(d_j;d_{<j})a_{\theta,t}(d_{<t})\notag\\
&\qquad\cdot\prod_{j<t}\frac{m_{\theta,j}(d_j;d_{<j})}{q_{\theta,j}(d_j;d_{<j})}\notag\\
&\quad=\sum_{t=1}^{K}\mathbb{E}_{D\sim Q_\theta}\!\left[a_{\theta,t}(D_{<t})\prod_{j<t}\alpha_{\theta,j}(D_j)\right]=J(\theta;c).
\label{eq:dreal_cancellation}
\end{align}%
}
Explicitly writing this prefix dependence makes the cancellation transparent: each $b_j(d_j; d_{<j})$ in the proposal distribution cancels with its denominator independently, regardless of how $a_{\theta,t}$ or $m_{\theta,j}$ depend on earlier tokens. After cancellation, multiplying and dividing by $q_{\theta,j}(d_j; d_{<j})$ recovers the $Q_\theta$-expectation, confirming $\widetilde J$ as an unbiased surrogate for any $\theta$, not just at $\theta = \bar\theta$. Since $Q_{\bar\theta}$ is fixed during backpropagation, differentiating the finite sum ensures that:
\begin{equation}
\mathbb{E}_{D\sim Q_{\bar\theta}}\left[\left.\nabla_\theta\widehat{\mathcal L}_{\mathrm{D\text{-}REAL}}\right|_{\theta=\bar\theta}\right]
= -\left.\frac{1}{K}\nabla_\theta J(\theta;c)\right|_{\theta=\bar\theta}.
\label{eq:dreal_unbiased_gradient}
\end{equation}

Hence, $\widehat{\mathcal L}_{\mathrm{D\text{-}REAL}}$ yields an unbiased stochastic gradient estimator (or subgradient, at the measure-zero set where $p_t(v) = q_{\theta,t}(v)$ for some $v$) for the normalized expected acceptance length. Applying the product rule to $\widetilde r_{\theta,t} = \alpha_{\theta,t}(D_t) \frac{q_{\theta,t}(D_t)}{\operatorname{sg}[b_t(D_t)]}$ for $p_t(D_t) \neq q_{\theta,t}(D_t)$ (at the non-differentiable kink $p_t(D_t) = q_{\theta,t}(D_t)$ the subgradient is a closed convex set containing both branches) provides an intuitive decomposition:
\begin{equation}
\left.\nabla_\theta\widetilde r_{\theta,t}\right|_{\bar\theta} = \left.\nabla_\theta\alpha_{\theta,t}(D_t)\right|_{\bar\theta} + \alpha_{\bar\theta,t}(D_t) \left.\nabla_\theta\log q_{\theta,t}(D_t)\right|_{\bar\theta},
\label{eq:dreal_score_term}
\end{equation}
where the second term cleanly recovers the score function (policy gradient) of the proposal trajectory~\citep{williams1992reinforce} without requiring heuristic baseline subtraction or clipping. Note that when $q_{\bar\theta,t}(D_t) > p_t(D_t)$, we have $\alpha_{\bar\theta,t}(D_t) = p_t(D_t)/q_{\bar\theta,t}(D_t)$, and the two terms on the right-hand side of Eq.~\eqref{eq:dreal_score_term} exactly cancel: the first term $\nabla_\theta \alpha_{\theta,t}(D_t)\big|_{\bar\theta} = \nabla_\theta(p_t(D_t)/q_{\theta,t}(D_t))\big|_{\bar\theta}$ yields $-(p_t/q_{\bar\theta,t}^2)\nabla_\theta q_{\theta,t}(D_t)\big|_{\bar\theta} = -\alpha_{\bar\theta,t}(D_t)\nabla_\theta \log q_{\theta,t}(D_t)\big|_{\bar\theta}$, which negates the second term. Thus $\nabla_\theta \widetilde r_{\theta,t}\big|_{\bar\theta} = 0$ in this region, consistent with the fact that $\widetilde r_{\theta,t}$ reduces to the $\theta$-independent ratio $p_t(D_t)/\operatorname{sg}[b_t(D_t)]$ when the draft over-proposes. Conversely, when $q_{\bar\theta,t}(D_t) < p_t(D_t)$, $\alpha_{\bar\theta,t}(D_t) = 1$ and the first term vanishes, leaving only the score function $\nabla_\theta \log q_{\theta,t}(D_t)\big|_{\bar\theta}$---exactly the gradient signal that encourages the drafter to assign more mass to the sampled token.

\paragraph{Reference-prefix surrogates vs.\ true verification expectations.}
It is important to distinguish the role of stop-gradient operators from the choice of training objective. Consider the unsmoothed D-PACE construction on a fixed offline reference block $y_{1:K}$~\citep{wu2026dpace}. With reference inputs held fixed, let $s_j(\theta):=q_\theta(y_j\mid c,y_{<j})$ and define the confidence proxy $S(\theta) = \sum_{t=1}^{K}\prod_{i=1}^{t}s_i(\theta)$ with positional weights $w_j(\theta) = \sum_{t=j}^{K}\prod_{i=1}^{t}s_i(\theta)$. Prior to heuristic weight smoothing, the detached weighted cross-entropy satisfies
\begin{equation}
\nabla_\theta\!\left[-\sum_{j=1}^K\operatorname{sg}[w_j(\theta)]\log s_j(\theta)\right]
= -\nabla_\theta S(\theta).
\label{eq:dpace_unsmoothed_gradient}
\end{equation}
Equation~\eqref{eq:dpace_unsmoothed_gradient} shows that the detached weighted-cross-entropy construction reproduces the exact gradient of the fixed-reference surrogate $S(\theta)$ before asymmetric confidence smoothing. However, $S(\theta)$ itself differs fundamentally from the true expected acceptance length $J(\theta;c)$ in Eq.~\eqref{eq:dreal_expected_length}: it evaluates draft probabilities assigned to individual reference tokens along a static teacher-forced sequence rather than distributional overlaps $a_{\theta,t}$ along online draft-sampled trajectories $D_{1:K} \sim Q_\theta$. Conversely, within our exact verification objective, replacing the importance ratio product $\prod_{j<t}\widetilde r_{\theta,j}$ with its detached value $\prod_{j<t}\operatorname{sg}[\alpha_{\theta,j}(D_j)]$ would discard the entire derivative of the prefix survival factor---including both the acceptance-ratio and score-function terms in Eq.~\eqref{eq:dreal_score_term}. By combining online distributional overlap with stop-gradient behavior denominators, D-REAL eliminates surrogate mismatch while maintaining an unbiased gradient for $J(\theta;c)$.

\subsection{Practical Implementation}
\label{sec:implementation}
\begin{algorithm}[t]
\caption{D-REAL Post-Training Step}
\label{alg:dreal_step}
\begin{algorithmic}[1]
\REQUIRE Draft parameters $\theta$, frozen target $p$, anchor distribution $\rho$, proposal count $K$
\STATE Sample a minibatch of anchors $\mathcal C\sim\rho$.
\STATE Compute block hidden states $H_{1:K}(c)\gets\mathrm{ParallelBackbone}_{\theta}(c)$ for all $c\in\mathcal C$.
\STATE Sample trajectory $D_t\sim q_\theta(\cdot \mid H_t(c), D_{<t})$ for $t=1,\ldots,K$ autoregressively (no gradient tracked), recording proposal log-probabilities $\log b_t(D_t)$ alongside each sampled token.
\STATE Obtain target distributions $p_t\gets\operatorname{sg}[p(\cdot\mid c,D_{<t})]$ online with prefix KV reuse.
\STATE Recompute draft distributions $q_{\theta,t}\gets q_\theta(\cdot \mid H_t(c), D_{<t})$ with gradients, parallelizing across $t$ via teacher-forced prefix tokens $D_{<t}$ from Step~3 (checkpointing backbone activations for memory efficiency).
\STATE Backpropagate $g\gets\nabla_{\theta}\widehat{\mathcal L}_{\mathrm{D\text{-}REAL}}$, retaining gradients through $m$, $a$, and $\widetilde r$.
\STATE Update $\theta\gets\mathrm{OptimizerStep}(\theta,g)$.
\end{algorithmic}
\end{algorithm}

\paragraph{Online target verification via random anchors.}
Algorithm~\ref{alg:dreal_step} summarizes a single training step of D-REAL. A central implementation requirement of D-REAL is obtaining the exact verification distributions $p_t(v) = p(v \mid c, D_{<t})$ conditioned on tokens $D_{<t}$ sampled from the evolving drafter rather than offline ground-truth continuations. To achieve this, we adopt an online verification protocol over random anchors. During each training step, anchor positions $c$ are sampled uniformly along offline target-generated sequences. At each anchor, a candidate block $D_{1:K}$ is sampled from the parallel drafter $q_\theta$, and the frozen target model is invoked online to compute $p_t$ along the \emph{actual draft-sampled prefix} $D_{<t}$.

\paragraph{Two-pass draft evaluation.}
Steps~3 and~5 evaluate the draft model twice for different purposes. Step~3 samples the draft trajectory $D_{1:K}$ autoregressively without tracking gradients, since discrete sampling is non-differentiable and storing backbone activations through the sequential loop would be prohibitively expensive. Step~5 then recomputes the draft distributions $q_{\theta,t}$ with gradients, treating the sampled tokens $D_{<t}$ as fixed teacher-forced inputs; because all prefix tokens are already known, the lightweight conditional head can evaluate all positions in parallel rather than sequentially, and activation checkpointing on the backbone hidden states trades compute for memory during backpropagation. The recorded proposal probabilities $b_t(D_t) = q_{\bar\theta,t}(D_t)$ from Step~3 serve as the stop-gradient denominators in Eq.~\eqref{eq:dreal_ratio}.

\paragraph{Numerical stability and variance control.}
All acceptance ratios, distributional overlaps, and prefix products are computed in $\mathtt{float32}$ to reduce numerical error in overlaps and the $K$-step product chain. The proposal probability $b_t(D_t) = q_{\bar\theta,t}(D_t)$ in the denominator of $\widetilde r_{\theta,t}$ is recorded at sampling time (with a small $\epsilon=10^{-8}$ floor for numerical safety) and detached via $\operatorname{sg}[\cdot]$. Notably, despite dividing by $b_t(D_t)$, D-REAL remains stable for updates without requiring PPO-style ratio clipping or heuristic reward baselines. This stability stems from two structural properties: 
(1)~\emph{Bounded on-policy ratios and local logit derivatives}: because $m_{\theta,t}(D_t) = \min\{p_t(D_t), q_{\theta,t}(D_t)\} \le q_{\theta,t}(D_t)$, evaluating $\widetilde r_{\theta,t}$ at $\theta=\bar\theta$ strictly bounds every prefix factor to $\alpha_{\bar\theta,t}(D_t) \in [0, 1]$ even when $b_t(D_t)$ is small. As shown in the discussion following Eq.~\eqref{eq:dreal_score_term}, the two gradient terms exactly cancel when $q_{\bar\theta,t}(D_t) > p_t(D_t)$, and only the log-likelihood gradient $\nabla_\theta \log q_{\theta,t}(D_t)$ survives when $q_{\bar\theta,t}(D_t) < p_t(D_t)$, completely avoiding $1/b_t(D_t)$ gradient explosion in logit space; and 
(2)~\emph{Rao-Blackwellization}: analytically marginalizing the step-$t$ token expectation into the exact overlap $a_{\theta,t}$ eliminates token-level sampling noise at the target position.

\paragraph{System efficiency and training overhead.}
Replacing teacher-forced offline continuations with prefixes sampled from the evolving drafter introduces two potential costs at every training step: an autoregressive draft rollout and online target verification of the sampled trajectories. The former is inexpensive: since parallel drafters produce all block hidden states in a single forward pass, trajectory sampling reuses that same pass and unrolls only the lightweight conditional head for $K$ steps under \texttt{no\_grad}, with sampled EOS tokens truncating the block early. The dominant added cost is therefore the online verification induced by the sampled prefixes, which we bound through three system-level optimizations using SGLang~\citep{zheng2024sglang} as the verification backend:
(1)~\emph{Offline prefix reuse}: the offline capture's target hidden state at a position already defines the exact verification distribution whenever the sampled prefix matches the reference continuation, so online scoring is invoked only from the first divergent token onward---at most $K{-}1$ positions per anchor, with no bonus-token evaluation;
(2)~\emph{Batched suffix scoring with prefix KV reuse}: for divergent anchors, RadixAttention shares the page-aligned KV states of the anchor context $c$, and the remaining short suffixes across anchors are grouped into a single batched extend forward; and
(3)~\emph{Detached target execution}: verification runs without gradient tracking and returns only final hidden states, deferring the vocabulary projection to the frozen shared LM head on the draft side---the target performs neither a backward pass nor any logit computation.
Consequently, the wall-clock overhead of unbiased prefix optimization remains modest: on an 8-GPU Hopper node, one epoch of D-REAL post-training on $1.3$M prompts requires approximately $2\times$ the wall-clock time of one baseline epoch.

\section{Experiment Setup}
\label{sec:exp_setup}
\paragraph{Models and Datasets.} 
We evaluate our method on two 5-layer parallel drafters, DSpark~\citep{cheng2026dspark} and DFlash2~\citep{inco2026dflash2}, using Qwen3-8B~\citep{qwen2025qwen3} as the frozen target model. Both drafters use a 5-layer block-parallel backbone that produces all block hidden states in a single forward pass conditioned on fused multi-layer target features, differing mainly in how intra-block token dependencies are modeled at the output head.

DSpark applies a low-rank (rank-$256$) Markov logit bias over the shared LM head conditioned on the previously drafted token, accompanied by an acceptance-confidence predictor, whereas DFlash2 uses a rank-$256$ candidate path selector to re-rank the top-$16$ unary candidates based on the predecessor token and draft hidden state. 

Both drafters are configured to propose $K=7$ speculative tokens per step. To construct the training corpus, we filter the full OpenPerfectBlend dataset~\citep{labonne2024openperfectblend} ($1.4$M prompts) to retain approximately $1.3$M valid prompts, and regenerate all responses offline using the Qwen3-8B target model in non-thinking mode (\texttt{enable\_thinking=False}) with $T=1.0$. Both drafters are trained following their publicly available reference recipes.

\paragraph{Implementation.} 
Our training pipeline is implemented in the SpecForge framework~\citep{li2026specforge} and executed on an 8-GPU NVIDIA Hopper node with a per-device batch size of $4$ (no gradient accumulation), a maximum sequence length of $4096$, and $512$ randomly sampled anchor positions per sequence. To evaluate post-training effectiveness on drafters whose validation acceptance length has largely plateaued, we adopt a two-stage protocol:
\begin{itemize}
    \item \textbf{Stage 1 (Baseline Pretraining):} We first train both drafters from scratch for $6$ epochs using the AdamW optimizer~\citep{loshchilov2019adamw} with a peak learning rate of $6\times 10^{-4}$, by which point the acceptance length has largely plateaued.
    \item \textbf{Stage 2 (Acceptance-Aware Post-Training):} Starting from the 6-epoch baseline checkpoint, we perform $1$ additional epoch of post-training on the same dataset using a reduced learning rate of $2\times 10^{-4}$ across all compared objectives.
\end{itemize}
During training, both the target and draft distributions are computed using unscaled logits ($T = 1.0$), and the resulting checkpoints are evaluated across both greedy ($T = 0.0$) and stochastic ($T = 1.0$) decoding at inference time.

\paragraph{Compared Methods.} 
We compare three training configurations under identical architectural and hyperparameter settings:
\begin{itemize}
    \item \textbf{Baseline:} The pretrained drafter optimized with the standard token-level objective and static exponential positional decay $w_t = \exp(-(t-1)/\gamma)$ with $\gamma = 4$ following~\citet{chen2026dflash}. We also evaluate a step-matched \textit{Baseline +1 epoch} trained for a 7th epoch under this original loss.
    \item \textbf{D-PACE~\citep{wu2026dpace}:} A reference-prefix weighted cross-entropy objective whose position weights incorporate cumulative confidence and continuation value. We apply confidence smoothing $\widetilde{s}_j=(1-\lambda)s_j+\lambda$ with $\lambda=0.5$ following the best practice of the original recipe when computing these weights. We evaluate D-PACE under the same one-epoch post-training protocol used for D-REAL.
    \item \textbf{D-REAL (Ours):} Post-training with our unbiased expected acceptance-length surrogate loss defined in Eq.~\eqref{eq:dreal_loss}, which preserves gradients through both the token-level overlap and the proposal trajectory distribution.
\end{itemize}

\paragraph{Benchmarks and Metrics.} 
We evaluate speculative decoding across $10$ benchmarks spanning three domains: \textit{Math} (GSM8K~\citep{cobbe2021gsm8k}, MATH-500~\citep{hendrycks2021math,lightman2024lets}, AIME25~\citep{maa2025aime}), \textit{Code} (HumanEval~\citep{chen2021humaneval}, MBPP~\citep{austin2021mbpp}, LiveCodeBench~\citep{jain2025livecodebench}), and \textit{Chat} (C-Eval~\citep{huang2023ceval}, MT-Bench~\citep{zheng2023mtbench}, Alpaca~\citep{taori2023alpaca}, Arena-Hard~\citep{li2025arenahard}). 

Across all tasks, models generate up to $2048$ tokens in non-thinking mode under both greedy ($T=0.0$) and stochastic ($T=1.0$) decoding, and we report the mean acceptance length $\tau$ (Eq.~\eqref{eq:emitted_length}) as the evaluation metric.

\definecolor{myhighlight}{RGB}{238, 246, 238} 

\begin{table*}[t]
\centering
\caption{Mean acceptance length ($\tau$) across ten benchmarks, including one additional verifier-generated token per round. \textbf{Bold} indicates the best results. \textbf{D-REAL (Ours)} consistently outperforms the baseline and D-PACE, particularly under stochastic sampling ($T=1.0$).}
\label{tab:acceptance}
\resizebox{\textwidth}{!}{%
\small
\setlength{\tabcolsep}{2.6pt}
\begin{tabular}{@{}llccccccccccc@{}}
\toprule
\multirow{2}{*}{\textbf{Model}} & \multirow{2}{*}{\textbf{Method}} & \multicolumn{3}{c}{\textbf{MATH}} & \multicolumn{3}{c}{\textbf{CODE}} & \multicolumn{4}{c}{\textbf{Chat}} & \multirow{2}{*}{\textbf{Avg.}} \\
\cmidrule(lr){3-5} \cmidrule(lr){6-8} \cmidrule(lr){9-12}
& & GSM8K & MATH-500 & AIME25 & HumanEval & MBPP & LCB & C-Eval & MT-Bench & Alpaca & Arena-Hard & \\
\midrule

\multicolumn{13}{c}{\textit{$T = 0.0$}} \\
\midrule
 & Baseline  & 6.43 & 6.39 & 5.82 & 5.82 & 5.45 & 5.78 & 3.69 & 3.84 & 3.75 & 3.77 & 5.07 \\
 & Baseline +1 epoch & 6.44 & 6.41 & 5.80 & 5.87 & 5.47 & 5.79 & 3.68 & \textbf{3.85} & 3.78 & 3.77 & 5.09 \\
 & D-PACE    & 6.47 & 6.41 & 5.82 & \textbf{5.88} & 5.47 & 5.80 & 3.69 & \textbf{3.85} & 3.78 & 3.77 & 5.09 \\
\rowcolor{myhighlight} \cellcolor{white}\multirow{-4}{*}{\textbf{DSpark}} 
 & \textbf{D-REAL} & \textbf{6.50} & \textbf{6.45} & \textbf{5.85} & \textbf{5.88} & \textbf{5.50} & \textbf{5.82} & \textbf{3.77} & 3.84 & \textbf{3.82} & \textbf{3.83} & \textbf{5.13} \\
\cmidrule{2-13}
 & Baseline  & 6.27 & 6.20 & 5.61 & 5.59 & 5.21 & 5.57 & 3.46 & 3.67 & 3.58 & 3.60 & 4.88 \\
 & Baseline +1 epoch & 6.28 & 6.19 & 5.64 & 5.63 & 5.22 & 5.62 & 3.45 & 3.69 & 3.59 & 3.60 & 4.89 \\
 & D-PACE    & 6.37 & 6.29 & 5.77 & 5.71 & 5.30 & 5.68 & 3.57 & 3.76 & 3.68 & 3.72 & 4.98 \\
\rowcolor{myhighlight} \cellcolor{white}\multirow{-4}{*}{\textbf{DFlash2}} 
 & \textbf{D-REAL} & \textbf{6.52} & \textbf{6.39} & \textbf{5.85} & \textbf{5.88} & \textbf{5.47} & \textbf{5.77} & \textbf{3.66} & \textbf{3.86} & \textbf{3.78} & \textbf{3.80} & \textbf{5.10} \\
\midrule

\multicolumn{13}{c}{\textit{$T = 1.0$}} \\
\midrule
 & Baseline  & 6.18 & 5.82 & 5.12 & 5.44 & 5.18 & 5.14 & 3.21 & 3.77 & 3.58 & \textbf{3.29} & 4.67 \\
 & Baseline +1 epoch & 6.20 & 5.82 & 5.07 & 5.45 & 5.22 & 5.08 & 3.24 & 3.76 & \textbf{3.62} & 3.28 & 4.67 \\
 & D-PACE    & 6.14 & 5.74 & 4.85 & 5.47 & 5.12 & \textbf{5.16} & 3.18 & 3.64 & 3.53 & 3.21 & 4.60 \\
\rowcolor{myhighlight} \cellcolor{white}\multirow{-4}{*}{\textbf{DSpark}} 
 & \textbf{D-REAL} & \textbf{6.30} & \textbf{5.90} & \textbf{5.20} & \textbf{5.55} & \textbf{5.27} & 5.15 & \textbf{3.31} & \textbf{3.79} & \textbf{3.62} & \textbf{3.29} & \textbf{4.74} \\
\cmidrule{2-13}
 & Baseline  & 6.16 & 5.93 & 5.39 & 5.38 & 5.05 & 5.13 & 3.21 & 3.72 & 3.63 & 3.34 & 4.69 \\
 & Baseline +1 epoch & 6.16 & 5.96 & 5.32 & 5.45 & 5.16 & 5.24 & 3.22 & 3.69 & 3.68 & 3.33 & 4.72 \\
 & D-PACE    & 6.01 & 5.73 & 4.97 & 5.23 & 5.02 & 5.04 & 3.26 & 3.77 & 3.61 & 3.32 & 4.60 \\
\rowcolor{myhighlight} \cellcolor{white}\multirow{-4}{*}{\textbf{DFlash2}} 
 & \textbf{D-REAL} & \textbf{6.52} & \textbf{6.21} & \textbf{5.49} & \textbf{5.80} & \textbf{5.44} & \textbf{5.26} & \textbf{3.38} & \textbf{3.91} & \textbf{3.70} & \textbf{3.44} & \textbf{4.92} \\
\bottomrule
\end{tabular}%
}
\end{table*}

\section{Experiment Results}
\label{sec:exp_results}
\subsection{Acceptance Length Comparison}
Table~\ref{tab:acceptance} compares D-REAL against the pretrained baseline, a continued-training baseline (\textit{Baseline +1 epoch}), and D-PACE across ten benchmarks. Across all 40 evaluated settings (2 architectures $\times$ 2 temperatures $\times$ 10 tasks), D-REAL outperforms D-PACE in 37 cases, ties in 1, and is lower in 2 (by $0.01$ tokens), consistently improving the macro-average $\tau$ for both DSpark and DFlash2. The gains are particularly pronounced on DFlash2, where D-REAL boosts average acceptance length by $+0.22$ tokens at $T=0.0$ and $+0.23$ tokens at $T=1.0$ over the baseline. Notably, \textit{Baseline +1 epoch} yields minimal macro-average changes ($\Delta \le 0.03$ tokens) across all configurations, indicating that standard token-level training has largely plateaued and that D-REAL's gains stem from the post-training objective rather than additional training steps.

The advantage of directly optimizing the true verification objective is clear under stochastic decoding ($T=1.0$). In our one-epoch post-training setting, D-PACE's macro-average $\tau$ degrades relative to the pretrained checkpoint (from $4.67$ to $4.60$ on DSpark and from $4.69$ to $4.60$ on DFlash2), whereas D-REAL improves the corresponding averages by $+0.07$ and $+0.23$ tokens. This contrast aligns with our theoretical analysis: because stochastic sampling frequently diverges from single reference continuations and accepts tokens via the overlap $\min\{p_t, q_{\theta,t}\}$, optimizing a reference-prefix cross-entropy surrogate suffers from objective mismatch at $T=1.0$. Although D-REAL is derived for stochastic rejection sampling, the improvements observed at $T=0.0$ provide empirical evidence of transfer to greedy decoding without temperature-specific retraining.

\subsection{End-to-End Serving Performance}
Because D-REAL purely optimizes the draft model parameters without altering the model architecture or verification protocol, improvements in acceptance length ($\tau$) translate directly into wall-clock speedups and require no additional inference-time overhead. To verify that these gains hold under realistic, high-concurrency workloads, we deploy the Qwen3-8B target model with both DFlash2 and DSpark drafters in SGLang~\citep{zheng2024sglang} and benchmark end-to-end serving performance on GSM8K at a concurrency of $32$ (Table~\ref{tab:sglang_perf}). 

Consistent with the GSM8K acceptance-length gains in Table~\ref{tab:acceptance}, D-REAL improves serving efficiency across both architectures and decoding temperatures. On DFlash2, D-REAL boosts output token throughput by \textbf{9.4\%} at $T=0.0$ ($17{,}634 \to 19{,}288$ tok/s) and \textbf{6.0\%} at $T=1.0$ ($10{,}064 \to 10{,}668$ tok/s), while reducing mean end-to-end request latency by \textbf{7.4\%} and \textbf{5.9\%}, respectively. Similarly, on the stronger DSpark baseline, D-REAL increases output throughput by \textbf{5.7\%} at $T=0.0$ ($17{,}949 \to 18{,}972$ tok/s) and \textbf{2.0\%} at $T=1.0$ ($10{,}440 \to 10{,}646$ tok/s), cutting mean latency by \textbf{3.6\%} and \textbf{1.9\%}. These results confirm that D-REAL delivers tangible system-level speedups in production-grade inference engines even under compute-intensive batched execution.

\begin{table}[t]
\centering
\caption{End-to-end serving performance of \textbf{DFlash2} and \textbf{DSpark} on GSM8K in SGLang, measured on an 8-GPU Hopper node (concurrency $= 32$). \textbf{D-REAL} translates higher acceptance lengths into substantial throughput gains and latency reductions without modifying the inference architecture.}
\label{tab:sglang_perf}
\renewcommand{\arraystretch}{1.15}
\setlength{\tabcolsep}{3pt}
\resizebox{\columnwidth}{!}{%
\begin{tabular}{llcc}
\toprule
\multirow{2}{*}{\textbf{Model}} & \multirow{2}{*}{\textbf{Method}} & \textbf{Output Tput} $\uparrow$ & \textbf{Mean E2E Latency} $\downarrow$ \\
 & & \footnotesize{(tok/s)} & \footnotesize{(ms)} \\
\midrule

\multicolumn{4}{c}{\textit{$T = 0.0$}} \\
\midrule
 & Baseline & 17,634 & 458 \\
\rowcolor{myhighlight} \cellcolor{white}\multirow{-2}{*}{\textbf{DFlash2}} 
 & \textbf{D-REAL} & \textbf{19,288} {\footnotesize(\textbf{+9.4\%})} & \textbf{424} {\footnotesize(\textbf{-7.4\%})} \\
\cmidrule{2-4}
 & Baseline & 17,949 & 449 \\
\rowcolor{myhighlight} \cellcolor{white}\multirow{-2}{*}{\textbf{DSpark}} 
 & \textbf{D-REAL} & \textbf{18,972} {\footnotesize(\textbf{+5.7\%})} & \textbf{433} {\footnotesize(\textbf{-3.6\%})} \\
\midrule

\multicolumn{4}{c}{\textit{$T = 1.0$}} \\
\midrule
 & Baseline & 10,064 & 824 \\
\rowcolor{myhighlight} \cellcolor{white}\multirow{-2}{*}{\textbf{DFlash2}} 
 & \textbf{D-REAL} & \textbf{10,668} {\footnotesize(\textbf{+6.0\%})} & \textbf{775} {\footnotesize(\textbf{-5.9\%})} \\
\cmidrule{2-4}
 & Baseline & 10,440 & 792 \\
\rowcolor{myhighlight} \cellcolor{white}\multirow{-2}{*}{\textbf{DSpark}} 
 & \textbf{D-REAL} & \textbf{10,646} {\footnotesize(\textbf{+2.0\%})} & \textbf{777} {\footnotesize(\textbf{-1.9\%})} \\
\bottomrule
\end{tabular}%
}
\end{table}

\subsection{Ablation Study}
\label{sec:ablation}

\begin{table}[t]
\centering
\caption{Ablation study on \textbf{DSpark} across representative benchmarks from each domain: GSM8K (Math), MBPP (Code), and C-Eval (Chat). Replacing the static decay with \textbf{D-REAL}'s unbiased dynamic prefix weighting yields consistent gains across all domains and temperatures.}
\label{tab:ablation_dspark}
\renewcommand{\arraystretch}{1.15}
\setlength{\tabcolsep}{3pt}
\resizebox{\columnwidth}{!}{%
\begin{tabular}{@{}llcccc@{}}
\toprule
\textbf{Temp.} & \textbf{Post-Training Objective} & \textbf{GSM8K} & \textbf{MBPP} & \textbf{C-Eval} & \textbf{Avg.} \\
\midrule
\multirow{4}{*}{$T=0.0$} 
 & Baseline (6 epochs)      & 6.43 & 5.45 & 3.69 & 5.19 \\
 & +1 Ep. Baseline Loss     & 6.44 & 5.47 & 3.68 & 5.20 \\
 & +1 Ep. TV + Static Decay & 6.48 & 5.49 & 3.72 & 5.23 \\
 \rowcolor{myhighlight} \cellcolor{white}
 & \textbf{+1 Ep. D-REAL (Ours)} & \textbf{6.50} & \textbf{5.50} & \textbf{3.77} & \textbf{5.26} \\
\midrule
\multirow{4}{*}{$T=1.0$} 
 & Baseline (6 epochs)      & 6.18 & 5.18 & 3.21 & 4.86 \\
 & +1 Ep. Baseline Loss     & 6.20 & 5.22 & 3.24 & 4.89 \\
 & +1 Ep. TV + Static Decay & 6.22 & 5.25 & 3.25 & 4.91 \\
 \rowcolor{myhighlight} \cellcolor{white}
 & \textbf{+1 Ep. D-REAL (Ours)} & \textbf{6.30} & \textbf{5.27} & \textbf{3.31} & \textbf{4.96} \\
\bottomrule
\end{tabular}%
}
\end{table}

To disentangle the impact of token-level distributional overlap from our trajectory-level unbiased prefix weighting, we evaluate an intermediate ablation variant on DSpark: \textbf{TV + Static Decay}. This variant optimizes local distributional overlap on online draft-sampled prefixes while replacing D-REAL's dynamic prefix product $\prod_{j<t}\widetilde r_{\theta,j}$ with the static exponential decay $w_t = \exp(-(t-1)/\gamma)$, yielding $\widehat{\mathcal{L}}_{\mathrm{TV+Static}} = 1 - \frac{1}{K}\sum_{t=1}^{K} w_t a_{\theta,t}$ (with $\gamma=4$). We ablate on DSpark as the weaker-gain case; the trend on DFlash2 is qualitatively similar.

Table~\ref{tab:ablation_dspark} reports the results on representative benchmarks from each domain (GSM8K for Math, MBPP for Code, and C-Eval for Chat). We observe a clear two-step improvement:
\begin{itemize}
    \item \textbf{Local TV overlap vs.\ Baseline loss:} Switching from the standard token-level loss (\textit{+1 Ep.\ Baseline Loss}) to online total variation overlap (\textit{TV + Static Decay}) improves the average acceptance length from $5.20 \to 5.23$ at $T=0.0$ and $4.89 \to 4.91$ at $T=1.0$, confirming the benefit of aligning the single-token objective with the exact verification acceptance rate.
    \item \textbf{Unbiased dynamic prefix vs.\ Static decay:} Incorporating D-REAL's differentiable survival prefix ($\prod_{j<t}\widetilde r_{\theta,j}$) brings a further, consistent boost across all three domains, raising the average $\tau$ to \textbf{5.26} ($T=0.0$) and \textbf{4.96} ($T=1.0$). Notably, the gain from dynamic prefix weighting is most pronounced under stochastic decoding ($T=1.0$, $+0.05$ on average and $+0.08$ on GSM8K), where sampling variance causes actual trajectory survival rates to deviate significantly from any fixed exponential schedule.
\end{itemize}

\begin{figure*}[!t]
    \centering
    \subfigure[Greedy Decoding ($T = 0.0$)]{%
        \includegraphics[width=\textwidth]{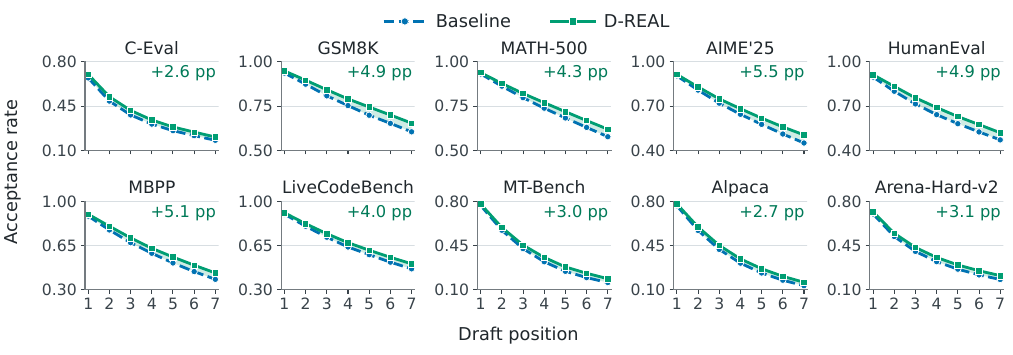}%
        \label{fig:accept_rate_t=0}%
    }
    \medskip
    \subfigure[Stochastic Decoding ($T = 1.0$)]{%
        \includegraphics[width=\textwidth]{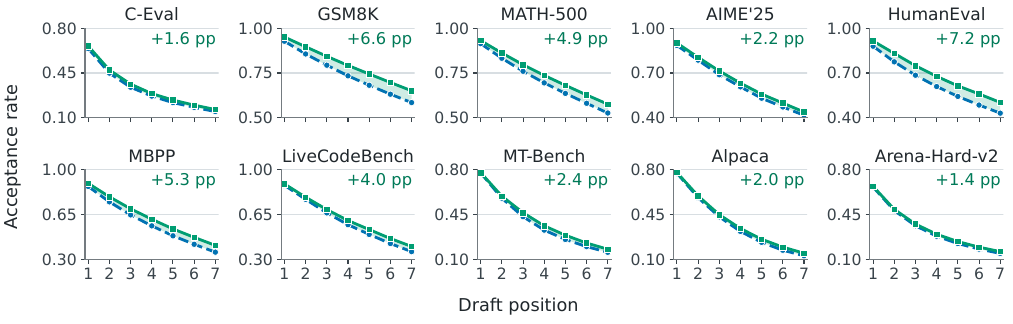}%
        \label{fig:accept_rate_t=1}%
    }
    \caption{Per-task cumulative acceptance rate $\Pr(N \ge t)$ across speculative positions $t \in \{1, \dots, 7\}$ for \textbf{DFlash2} under (a) greedy ($T=0.0$) and (b) stochastic ($T=1.0$) decoding. Across both regimes and all ten benchmarks, \textbf{D-REAL} consistently slows down prefix survival decay compared to the baseline, yielding compounding gains with draft depth. Y-axis ranges vary by workload and are shared across temperatures; shading highlights the gap, and labels report the position-7 gain in percentage points.}
    \label{fig:accept_rate_all}
\end{figure*}

\subsection{Position-wise Prefix Survival Analysis}
Recall from Eq.~\eqref{eq:dreal_expected_length} that the expected acceptance length decomposes into the sum of prefix survival probabilities across speculative steps, $\mathbb{E}[N \mid c] = \sum_{t=1}^K \Pr(N \ge t \mid c)$. To verify whether D-REAL genuinely improves multi-step trajectory survival rather than merely boosting first-token accuracy, Table~\ref{tab:pos_accept} compares the macro-averaged cumulative acceptance rate $\Pr(N \ge t)$ across all seven draft positions ($t \in \{1, \dots, 7\}$) on DFlash2, with per-task survival curves detailed in Figures~\ref{fig:accept_rate_t=0} and~\ref{fig:accept_rate_t=1}.

\begin{table}[!htbp]
\centering
\caption{Macro-averaged cumulative acceptance rate $\Pr(N \ge t)$ (\%) across speculative positions $t \in \{1, \dots, 7\}$ for \textbf{DFlash2} over all ten benchmarks  ($\tau \approx 1 + \sum_{t=1}^7 \Pr(N \ge t)$, where the small systematic positive bias stems from early-EOS block truncation causing position-$t$ rates to be computed conditional on the block reaching $t$). While D-PACE drops below the baseline at $T=1.0$, \textbf{D-REAL} achieves consistent, substantial gains across the entire draft depth.}
\label{tab:pos_accept}
\renewcommand{\arraystretch}{1.12}
\small
\setlength{\tabcolsep}{1.8pt}
\resizebox{\columnwidth}{!}{%
\begin{tabular}{@{}llrrrrrrr@{}}
\toprule
\textbf{Temp.} & \textbf{Method} & \textbf{1} & \textbf{2} & \textbf{3} & \textbf{4} & \textbf{5} & \textbf{6} & \textbf{7} \\
\midrule
 & Baseline & 83.6 & 70.7 & 60.5 & 52.6 & 46.2 & 40.8 & 35.9 \\
 & D-PACE   & 84.2 & 71.8 & 62.1 & 54.4 & 48.1 & 42.7 & 37.9 \\
 \rowcolor{myhighlight} \cellcolor{white}
 & \textbf{D-REAL} & \textbf{85.0} & \textbf{73.2} & \textbf{63.7} & \textbf{56.2} & \textbf{50.0} & \textbf{44.8} & \textbf{39.9} \\
 \cmidrule(lr){2-9}
 \multirow{-4}{*}{$T=0.0$} & \textit{$\Delta$ vs. Base} & +1.4pp & +2.5pp & +3.2pp & +3.6pp & +3.8pp & +4.0pp & +4.0pp \\
\midrule

 & Baseline & 82.1 & 68.6 & 58.1 & 49.8 & 43.1 & 37.5 & 32.6 \\
 & D-PACE   & 80.8 & 67.0 & 56.3 & 48.3 & 41.7 & 36.3 & 31.7 \\
 \rowcolor{myhighlight} \cellcolor{white}
 & \textbf{D-REAL} & \textbf{83.7} & \textbf{71.3} & \textbf{61.4} & \textbf{53.5} & \textbf{47.0} & \textbf{41.4} & \textbf{36.4} \\
 \cmidrule(lr){2-9}
 \multirow{-4}{*}{$T=1.0$} & \textit{$\Delta$ vs. Base} & +1.6pp & +2.7pp & +3.3pp & +3.7pp & +3.9pp & +3.9pp & +3.8pp \\
\bottomrule
\end{tabular}%
}
\end{table}

Table~\ref{tab:pos_accept} summarizes two empirical observations. First, under our post-training protocol, D-PACE improves macro-averaged survival probabilities over the baseline at $T=0.0$ but falls below it at every position for $T=1.0$. D-REAL improves these probabilities at all seven positions under both evaluated temperatures. Second, D-REAL maintains strong gains deep into the speculative block: rather than decaying in later steps, the survival margin over the baseline grows from $+1.4$ percentage points at $t=1$ to $+4.0$ points at $t=7$ under $T=0.0$, and from $+1.6$ points at $t=1$ to a peak of $+3.9$ points at $t=5,6$ ($+3.8$ points at $t=7$) under $T=1.0$, corresponding to over \textbf{+11\%} relative improvement at the final speculative token.

The per-task survival curves in Figures~\ref{fig:accept_rate_t=0} and~\ref{fig:accept_rate_t=1} confirm that this compounding advantage holds uniformly across all ten benchmarks rather than being driven by a few outlier tasks. Under both greedy ($T=0.0$) and stochastic ($T=1.0$) decoding, \textbf{D-REAL} (green curves) lies strictly above the baseline across every single task and speculative position, with the vertical gap between the two curves progressively widening from $t=1$ to $t=7$. This widening gap is especially pronounced under stochastic verification ($T=1.0$, Figure~\ref{fig:accept_rate_t=1}): on structured reasoning and coding tasks such as GSM8K, MATH-500, HumanEval, and MBPP, D-REAL substantially flattens the survival decay slope, elevating the final-position ($t=7$) acceptance rate by roughly $5$--$7$ percentage points and retaining $45\%$--$65\%$ survival at the end of the block. Even on higher-entropy benchmarks (e.g., C-Eval, MT-Bench, and Arena-Hard), D-REAL consistently delays trajectory rejection and maintains a clear positive margin through the deepest positions. Together, these results confirm that optimizing our prefix-weighted objective (Eq.~\eqref{eq:dreal_surrogate}) enables parallel drafters to sustain higher target agreement across the full speculative horizon.

\FloatBarrier
\section{Limitations and Future Work}
\label{sec:limitations}

\paragraph{Training Overhead.} 
Unlike offline token-level objectives that rely solely on pre-computed target trajectories, D-REAL requires sampling draft prefixes and querying the target model online to obtain exact verification distributions along draft-proposed trajectories. Despite our system-level optimizations---including offline prefix reuse, RadixAttention KV sharing, and detached target execution (Section~\ref{sec:implementation})---this online verification step inevitably introduces additional per-step training latency compared to standard teacher-forced training. Nevertheless, because D-REAL is applied as a lightweight, single-epoch post-training stage and introduces zero runtime overhead during deployment, this one-time training cost is rapidly amortized in production serving. Exploring asynchronous target scoring or truncated verification horizons to further reduce this training overhead remains a practical direction for future systems optimization.

\paragraph{From Random Anchors to On-Policy Contexts.} 
While D-REAL eliminates gradient bias for optimizing the expected acceptance length conditioned on a given context $c$, a remaining distribution mismatch lies in how these training contexts $c \sim \rho$ are collected. Following standard practice in parallel drafter training, we sample anchors uniformly along offline target-generated sequences. However, during actual speculative deployment, the sequence of verification-round boundaries (i.e., the exact positions where drafting begins) is dynamically determined by the acceptance lengths of preceding rounds. A natural direction for future work is to extend D-REAL from random offline anchors to fully on-policy speculative rollouts or replay buffers, thereby aligning the training anchor distribution $\rho$ with the exact state visitation distribution encountered at inference time.

\section{Conclusion}
In this paper, we presented D-REAL, a principled post-training objective designed to resolve the objective mismatch between token-level draft training and sequential speculative verification. By preserving the dependencies of verification contexts on draft-sampled prefixes and properly accounting for the evolving proposal distribution, D-REAL provides an unbiased gradient estimator for the single-round expected acceptance length. Our derivation reveals that combining pointwise distributional overlap with stop-gradient proposal ratios cleanly captures the policy gradient of the draft trajectory without introducing optimization heuristics. Extensive experiments on DSpark and DFlash2 across ten diverse benchmarks demonstrate that one epoch of D-REAL post-training consistently unlocks further acceptance-length gains from baselines whose validation acceptance length has largely plateaued under the original training recipe, surpassing both continued pretraining and D-PACE across stochastic and greedy decoding. As speculative decoding increasingly shifts toward high-efficiency parallel drafters, D-REAL establishes a solid theoretical and practical verification-aligned post-training paradigm for speculative decoding.

\clearpage
\bibliography{references}
\bibliographystyle{mlsys2025}

\end{document}